# An Agentic Framework Using Rules and LLMs for Embedding and Annotating Descriptive Document Layouts: A Plant Science Use Case

Nicolas Turenne(1)(2), Eric Chenin(2), Youcef Sklab(2) & Jean-Daniel Zucker(2)

Affiliations

1. INRAE, MathNum, France.

2. IRD, Sorbonne Université, UMMISCO, Paris France.

corresponding author(s): Nicolas Turenne (Email: nicolas.turenne@ird.fr)

## Abstract

Background: Recent advances in information retrieval (IR) leverage both dense and sparse representations, large language models (LLMs), and specialized retrieval models to improve ranking accuracy, relevance, and cross-lingual performance. Complementary techniques such as passage indexing, document layout analysis, and semantic knowledge representation further enhance retrieval effectiveness by capturing fine-grained contextual and structural information. Emerging agentic LLM frameworks extend these capabilities by enabling planning, iterative reasoning, tool use, and multi-agent collaboration, thereby broadening applications across diverse domains. These frameworks also emphasize rigorous evaluation, ethical considerations, and trustworthiness, ensuring responsible deployment in real-world settings.

We propose a modular, agent-based pipeline for botanical trait extraction. Optical character recognition (OCR) converts PDFs into machine-readable text, while segmentation and indexing organize content by genus and species. Rule-based parsers extract structured botanical traits, and ensembles of large language models (LLMs) expand trait vocabularies and resolve ambiguities. This approach ensures accurate species recognition, scalable annotation, and explainable integration of textual botanical descriptions, enabling robust and interpretable data extraction across large botanical corpora.

Results: Using three regional botanical datasets, our system extracted 55,737 trait annotations across 4,961 species, averaging 9.1 traits per species. Integration of LLM-based enrichment improved coverage for 75% of traits, increasing total annotations by 59%. While the choice of OCR engine had a minor effect on species recognition, overall annotation counts remained stable, demonstrating the robustness, scalability, and reliability of the pipeline for large-scale botanical trait extraction.

## 1 Introduction

In this paper, we present a novel approach for indexing named entities from descriptive text documents. By embedding domain knowledge into a searchable catalog index, our method provides a comprehensive and standardized resource for information retrieval and analysis. Our use case focuses on plant features—specifically morphological characteristics, habitat, life cycle, and geographic distribution—collectively referred to as plant traits by botanists and ecologists (Kattge et al., 2020). Taxonomic plant descriptions serve as a foundational source of knowledge for botany, ecology, conservation, and environmental science. Numerous international platforms, such as GBIF (2025) and World Flora Online (2025), offer extensive information on plant taxonomy and distribution, which our framework leverages to enhance coverage, consistency, and usability of trait data.World Flora Online is an ambitious project that aims to embed the botanical world by building a searchable catalog index of plant species descriptions (Miller. 2019). The project is collaborative, and the development of the WFO is a challenging task, but the end result will be a comprehensive and standardized resource for the global botanical community, providing basic information on all of the world's plants (Jackson et al. 2015 ; World Flora Online, 2025)

Recent advances in information retrieval (IR) integrate dense and sparse representations (Schütze et al., 2008)(Lopes, 2022), large language models (LLMs), and retrieval models (RMs) to enhance precision, ranking, and cross-lingual performance. Passage indexing, document layout analysis, and semantic knowledge representation further improve retrieval effectiveness. Scientific knowledge extraction relies on advanced IR and document vectorization techniques, combining dense and sparse embeddings, supervised and unsupervised learning, proximity models, and query expansion to optimize document relevance. Modern approaches leverage LLMs and RMs in frameworks such as InteR and Self-Retrieval, enabling zero-shot retrieval and hierarchical, context-aware search. Passage indexing has progressed from single-representation models, like ANCE, to multi-representation models, such as ColBERT, improving retrieval precision for complex queries, with fine-grained proposition-level indexing further supporting cross-task generalization (Khattab et al, 2020) (Macdonald et al. 2021). Document layout analysis integrates vision-based methods (e.g., YOLOv5, VILA) (Sugiharto et al. 2023) with graph-based approaches (e.g., Doc-GCN) to capture structural and semantic relationships, facilitating accurate indexing of text, tables, and multimedia components (Luo et al. 2022). Agentic frameworks extend LLMs with planning, memory, tool integration, and iterative reasoning, enabling autonomous, context-sensitive decision-making across domains including healthcare, finance, cybersecurity, and education. Multi-agent systems simulate emergent social behaviors, while agentic memory and unlearning frameworks improve adaptability and data management. Evaluation is critical, with benchmarks like BEIR, TREC-COVID, and BALROG measuring retrieval and agentic reasoning performance (Thakur et al., 2021)(Paglieri et al., 2025).

State-of-the-art approaches leverage neural retrieval, multimodal indexing, and agentic large language model (LLM) frameworks, combining planning, reasoning, and multi-agent collaboration to enable scalable, precise, and domain-adaptable knowledge extraction from complex scientific literature. While these methods emphasize efficiency, scalability, and general-purpose applicability, they often lack domain-specific ontological adaptation for specialized scientific corpora. To address this gap, we present an explainable, agent-based pipeline for botanical text mining that integrates OCR, segmentation, indexing, and rule-based parsers to support large-scale trait extraction. We further enhance botanical terminology through an iterative Object Class Learning (OCL) approach, leveraging an ensemble of LLMs to expand and refine vocabulary. Finally, we demonstrate the scalability and robustness of our framework through large-scale evaluations on regional floras from New Caledonia, Senegal, and Cameroon, validating its applicability across diverse botanical datasets.

These Key Points highlight originality and innovative aspects of our proposed method:

- A modular agentic architecture for ecological text mining — integrating OCR, document segmentation, and passage indexing with LLM-enhanced reasoning for large-scale, interpretable trait extraction from botanical corpora.
- Hybrid rule-based and LLM-driven trait annotation — combining deterministic linguistic patterns with iterative vocabulary enrichment from multiple LLMs to improve coverage and precision while maintaining transparency and explainability.

- Demonstration on global botanical datasets — applying the framework to thousands of species descriptions from regional floras (New Caledonia, Senegal, and Cameroon), achieving over 55,000 morphological trait annotations across 4,961 species.

## 2 State of the art

Scientific knowledge extraction from literature relies on effective document vectorization and representation techniques.

### 2.1 Information retrieval

Information retrieval (IR) is the field concerned with the representation, storage, organization, and access of information items, aiming to retrieve relevant content according to a user's query (Schütze et al., 2008; Lopes, 2022). IR encompasses diverse data types, including text, images, and videos. Traditional IR systems employ term proximity to improve ranking models beyond the "bag of words" assumption, promoting higher scores for documents where query terms appear close together (Giang et al., 2015). Proximity models can improve retrieval performance by 4–7% in terms of Mean Average Precision. Query expansion techniques further enhance retrieval by identifying additional documents relevant to user requirements (Silva et al., 2021).

IR systems can be characterized along two dimensions: dense versus sparse vector representations (Izacard, 2021) and supervised versus unsupervised approaches (Lin, 2021a; Lin, 2021b). Dense and sparse retrieval can be integrated in cross-lingual systems by combining proximity and original scores for document reranking (Giang et al., 2015). Document representation relies on vectorization, evolving from weighted-frequency models (Turenne, 2016) to fixed-size embeddings (Devlin, 2019). Recent advances leverage large language models (LLMs) and retrieval models (RMs) to enhance retrieval effectiveness (Nogueira & Lin, 2019; Devin et al., 2019; Zhu et al., 2023; Feng et al., 2023). Frameworks such as InteR combine LLMs and RMs to achieve superior zero-shot retrieval performance, while Self-Retrieval integrates indexing, retrieval, and reranking within a single LLM (Tang et al., 2024). Hierarchical retrieval mechanisms improve accuracy and clarify ambiguous semantics (Zhang et al., 2024), and sequence-generation-based frameworks enable universal and flexible information extraction (Lu et al., 2022).

Visual IR methods leverage graphical interfaces and optimized tag clouds to enhance usability and relevance (Hassan-Montero & Herrero-Solana, 2024). Integration with semantic web technologies and knowledge graphs supports structured, machine-interpretable representations for improved search, reasoning, and knowledge discovery (Martinez-Rodriguez & Hogan, 2020). Rigorous evaluation using benchmarks like BEIR (Thakur et al., 2021) and domain-specific collections such as TREC-COVID (Voorhees & Kanoulas, 2021) ensures reproducible performance assessment. Specialized applications, for example in NLP-driven materials research, rely on these techniques to maintain retrieval accuracy and reliability in real-world contexts (Olivetti et al., 2020).

### 2.2 Passage Indexing and Representation

(Bell, 2020) offers practical guidance on indexing biographical texts. It addresses unique challenges like chronological narratives, multiple identities, and complex relationships, providing strategies to create clear, accessible indexes that enhance readers' ability to navigate personal histories and life stories. Passage indexing is a crucial step in dense passage retrieval, where a large collection of passages is encoded and indexed for efficient retrieval (Macdonald et al. 2021). Two common approaches to passage indexing are single representation and multiple representations. Single representation approaches, such as ANCE, represent entire passages with a single embedding, usually the [CLS] token of a BERT model (Khattab et al, 2020)). These approaches are more efficient in terms of response time and memory usage but may underperform multiple representation approaches in terms of Mean Average Precision (MAP) and Mean Reciprocal Rank at 10 (MRR@10) (Macdonald et al. 2021). Multiple representation approaches, such as ColBERT, represent each token in a passage with its own embedding. These approaches are more effective than single representation approaches for MAP and MRR@10 and obtain better improvements than single representations for queries that are hard for Best Matching 25 (BM25), definitional queries, and those with complex information needs. Recent studies have shown that indexing the corpus at a finer-grained level, such as proposition level, can improve passage retrieval performance (Chen et al. 2023). This is because finer-grained indexing allows for more precise retrieval and can improve cross-task generalization. To improve passage retrieval performance, some approaches use contextualized exact term matching and efficient passage expansion. These approaches can improve ranking effectiveness without increasing query latency (Zhuang et al. 2021). Approaches propose a learning-based approach to estimate passage weighting within inverted indexes (Mallia et al, 2021), term weighting Mackenzie et al. (2020). Dai and Callan (2020) introduce DeepCT, a context-aware term weighting framework for first-stage passage retrieval. Other approaches use zero-shot question generation to re-score retrieved passages. This approach can be applied on top of any retrieval method and provides rich cross-attention between query and passage (Sachan et al. 2022). Li et al.

(2024) introduce PARADE, a Transformer-based model that aggregates passage-level relevance signals to predict document-level relevance scores. Leonhardt et al. (2024) present ‘Fast-Forward Indexes,’ a vector index structure leveraging dual-encoder models to efficiently re-rank large result sets. In addition, some studies have proposed automated frameworks for optimizing the retrieval augmented generation (RAG) pipeline (Kim et al. 2024). These frameworks can help improve the accuracy of retrieval modules and ensure efficient use of prompt tokens. Zhuang et al. (2022) propose DSI-QG, a refined Differentiable Search Index framework that addresses the mismatch between indexing long documents and retrieving via short queries (especially in cross-lingual scenarios).

### 2.3 Document layout analysis

Document layout indexing is a critical step in document analysis, involving the creation of a searchable index of a document’s structural components (Bansal et al., 2016; Bhowmik, 2023). This process enables efficient retrieval of specific layout features, supporting applications such as document search, layout-based retrieval, and automated analysis. Approaches to layout indexing generally fall into two categories: graph-based and vision-based methods.

Graph-based methods represent the document layout as a graph, where nodes correspond to layout components and edges represent their relationships. These graphs facilitate efficient indexing and retrieval. For example, Doc-GCN (Luo et al., 2022) uses graph convolutional networks to capture relationships between layout components, enhancing layout analysis performance. Vision-based methods apply computer vision techniques to detect layout elements. YOLOv5 (Sugiharto et al., 2023) identifies components such as paragraphs, tables, and images to generate an index of layout features. VILA (Shen et al., 2022) further improves structured content extraction from scientific PDFs by leveraging visual layout groups. Clustering-based methods, such as those proposed by Tomovic et al. (2021), align layouts extracted from multiple OCR engines.

Document layout indexing has been applied to diverse document types, including unstructured newspapers (Bansal et al., 2016) and PDF documents (Pfitzmann et al., 2022; Peña et al., 2023). Historical and scientific documents benefit from specialized approaches: anisotropic diffusion with geometric features for historical layouts (BinMakhashen & Mahmoud, 2023), comparative evaluations of tabular segmentation techniques (Liang et al., 2021), and comprehensive annotated datasets for scientific articles (Gemelli et al., 2024).

Recent methods integrate semantic information and relation modeling. VSR (Zhang et al., 2021) unifies vision, semantics, and relations, while PP-DocLayout (Sun et al., 2025) accelerates large-scale layout detection. Other approaches include multilingual document indexing incorporating page numbers (Nirmala & Lavanya, 2025), enhanced object detection for structural component segmentation (Minouei et al., 2021), multimodal pretraining for layout-aware understanding (Wu et al., 2021), and invoice analysis combining text and layout features (Ha & Horák, 2022).

### 2.4 Agentic framework and LLM

Botti (2025) critically examines agentic AI and multi-agent systems, questioning whether current developments merely reinvent established multi-agent principles. The study highlights conceptual overlaps, distinctions, and challenges, advocating for clearer definitions and integration of prior multi-agent research to advance the capabilities and practical applications of agentic AI. Plaat et al. (2025) provide a comprehensive survey of agentic large language models (LLMs), analyzing their architectures, decision-making mechanisms, and multi-agent interactions.

Agentic frameworks endow LLMs with structured mechanisms for planning, reasoning, tool invocation, and adaptation based on intermediate outcomes, enhancing their versatility in sustained, context-aware decision-making scenarios (Yao et al., 2025). These frameworks often involve layered or iterative LLM calls, extending the chain-of-thought paradigm to emulate autonomous agent-like behavior. Agentic LLMs can be specialized into domain-specific agents, for example, in programming synthesis or cardiological diagnostics (Liu et al., 2025). During inference, their performance can be further enhanced through structured prompts, reasoning regularization, or integration of external tools for problem-solving in specific tasks.

Recent innovations include ALU, an agentic multi-agent LLM unlearning framework that removes information at inference without retraining or accessing model weights, while preserving model utility and adapting to unlearning requests in real time (Sanyal & Mandal, 2025). QuBE (Question-based Belief Enhancement) improves context awareness in partially observable environments by constructing belief states through targeted question-answering (Kim et al., 2024). A-Mem introduces an agentic memory architecture inspired by the Zettelkasten method, updating contextual representations, generating semantic links, and evolving memory over time (Xu et al., 2025). Practical design considerations across planning, memory, tools, and control flow have been framed to guide real-world agentic LLM deployment (Sypherd & Belle, 2024).

Agentic LLMs demonstrate broad applicability: in financial question answering (Wang et al., 2025), cybersecurity and threat intelligence (Saha et al., 2025; Yao et al., 2025), education and essay scoring (Tirupathi et al., 2025), clinical reasoning (Das et al., 2024), and adaptive environmental decision-making (Dolant & Kumar, 2025). Multi-agent LLM systems can simulate emergent social behaviors, evaluate theoretical models at scale, and explore phenomena difficult to study empirically (Haase et al., 2025). Reflexion-based workflows, for example, have been applied to generate patient-friendly radiology reports, improving readability and comprehension (Sudarshan et al., 2024).

Evaluation remains a central concern. BALROG provides a benchmark for assessing agentic reasoning across long-horizon, multi-step tasks for LLMs and VLMs (Paglieri et al., 2025). Ethical, legal, and safety considerations are addressed through frameworks such as TRiSM for trust, risk, and security management (Raza et al., 2025) and Raptis et al.'s framework for fairness, transparency, robustness, safety, and oversight. Gabison and Xian (2025) further examine liability and compliance issues, highlighting challenges in deploying agentic AI within complex socio-technical environments. Overall, agentic LLMs offer a paradigm shift, enabling LLMs to operate more autonomously, effectively, and safely across diverse domains (Kamalov et al., 2025).

Key differences with our method:

The first key difference is botanical domain specialization. While state-of-the-art (SoTA) approaches focus broadly on general-purpose information retrieval (IR) and knowledge extraction, our method is domain-specific, designed for large-scale botanical corpora encompassing millions of species descriptions, herbarium labels, and Wikipedia entries. This specialization requires trait-specific semantic parsing, such as morphological feature extraction, which is not addressed in generic IR pipelines.

The second key difference is the hybrid NLP and ontology-based approach. Our framework combines rule-based gazetteers, fine-tuned named entity recognition (NER), and ontology-driven standardization, whereas SoTA methods rely primarily on dense and sparse embedding-based retrieval models. This hybrid strategy enhances precision in extracting structured plant traits and reduces noise arising from overlapping or ambiguous terminology.

## 3 Dataset and resources

### 3.1 Text dataset

More than 550 monographs were included in this study (Turenne et al., 2025), covering approximately 36 geographical regions. The Optical Character Recognition (OCR) process required 127 hours of computation. Most PDFs were originally published between 1845 and 2000 and consist of digitized scans of printed botanical volumes (Figure 1). Consequently, OCR was essential to convert these scanned pages into machine-readable text suitable for downstream natural language processing (NLP) applications.

We employed the open-source Tesseract OCR engine (version 5.5.0) to perform text recognition on the scanned documents. Tesseract is widely used for OCR tasks and supports multiple languages, making it well-suited for historical and low-quality scans. Version 5.5.0 integrates classical image processing algorithms with modern neural network techniques, notably Long Short-Term Memory (LSTM) recurrent networks optimized for sequential data. The LSTM model processes features extracted through convolutional layers, followed by sequential modeling in LSTM layers, and can optionally leverage integrated dictionaries and language models based on grammatical rules and frequency tables. This OCR step was essential for converting image-based PDFs into machine-readable text suitable for subsequent natural language processing (NLP) analyses.

We compared Tesseract with classical open-access PDF-to-text tools using a subset of 12,000 pages from New Caledonia, Cameroon, and Senegal. We observed a 32% increase in genus detection and a 14% increase in species name recognition. However, after manual annotation, no significant improvement in overall accuracy was observed. Despite this, we adopted Tesseract OCR for converting scanned PDFs into machine-readable text files due to its robustness and superior performance in preliminary extraction.

The corpus comprises documents in English, French, Spanish, Portuguese, Mandarin Chinese, and Russian.

### 3.2 Dictionaries

To structure the dataset, each text segment was organized into a file-species unit, where each file corresponds to a specific plant species and is named accordingly. Prior studies have highlighted the critical importance of accurately extracting scientific names from text (Akella, 2012), as species names serve as key identifiers linking disparate botanical knowledge sources. Unlike Nainia et al. (2024), who employed a fine-tuned language model for species name detection, we adopted a computational linguistics approach based on lexical transformations and curated dictionaries. This method is implemented in an algorithm we term *WordGen* (cf. methods section).

We merge three major botanical resources:

- World Checklist of Vascular Plants (WCVP, 2025),
- TRY Database (Kattge et al., 2020),
- World Flora Online (WFO, 2025).

These resources provide standardized scientific names along with metadata indicating whether a name corresponds to an accepted taxon or a synonym, at both genus and species levels. A genus groups multiple species, and the standard format for scientific names follows binomial nomenclature: genus epithet (e.g., *Quercus pubescens*). In cases where subspecies are identified (e.g., *Quercus pubescens subsp. anatolica*), we treat the subspecies as a synonym of its parent species for consistency and simplification; thus, *Quercus pubescens subsp. anatolica* is considered a synonym of *Quercus pubescens*.

Our compiled dictionary contains over 1.6 million entries encompassing species, subspecies, and synonyms. After normalizing subspecies and synonyms to their corresponding accepted species names, we retain approximately 1.2 million unique species-level names (see Table 1).

**Tab. 1** Number of genera and species in the merged botanical dictionaries

| scientific names (including synonyms) | merge | WCVP |
|---|---|---|
| Genera | **46,068** | 34,798 |
| Taxa | **1,127,959** | 996,093 |

The total number of accepted plant species names remains debated across taxonomic databases, with 388,342 listed in World Flora Online (WFO) and 407,961 in the World Checklist of Vascular Plants (WCVP) (Schellenberger Costa et al., 2023). Accurately identifying the beginning of each species section is critical for extracting meaningful, species-specific information. This task is particularly challenging due to textual inconsistencies and OCR-induced errors in digitized documents. Consequently, the development of a robust species name recognition tool is essential.

**6. MACADAMIA** F. von Mueller

Trans. Phil. Inst. Vict., **2**, 1858, p. 72.

Arbres ou arbustes élevés.

Feuilles simples, entières ou ± dentées, rapprochées en faux-verticilles, subopposées ou alternes.

Inflorescences en grappes simples ou composées, multiflores, terminales, subterminales, axillaires, pseudo-axillaires ou latérales et, dans ce dernier cas, naissant souvent sur le vieux bois.

**2. Macadamia Rousselii** (Vieillard) Sleumer

Sleumer, Blumea, **8**, 1955, p. 4.
— *Rhopala (Roupala) Rousselii* Vieill., Bull. Soc. Linn. Normandie, **9**, 1865, p. 394.

Arbre pouvant atteindre 12 m de hauteur. Rameaux arrondis. Écorce brunâtre ou grisâtre, rugueuse ou lisse, présentant des lenticelles ± nombreuses.

— 115 —

**Fig. 1** Sections corresponding to species within a taxonomic volume. As illustrated, *Pteris tripartita* and *Pteris novae-caledoniae* correspond to distinct blocks of descriptive content (left). Each species block begins with a section label indicating the genus name (right)

## 4 Methods

### 4.1 General architecture

Our agent architecture processes classification data by indexing descriptive files, segmenting pages, and classifying them using hierarchical categories (see Figure 2). Annotation is performed based on recognized entity names. This pipeline ensures efficient mapping, accurate classification, and streamlined preparation of data for downstream categorization analyses.

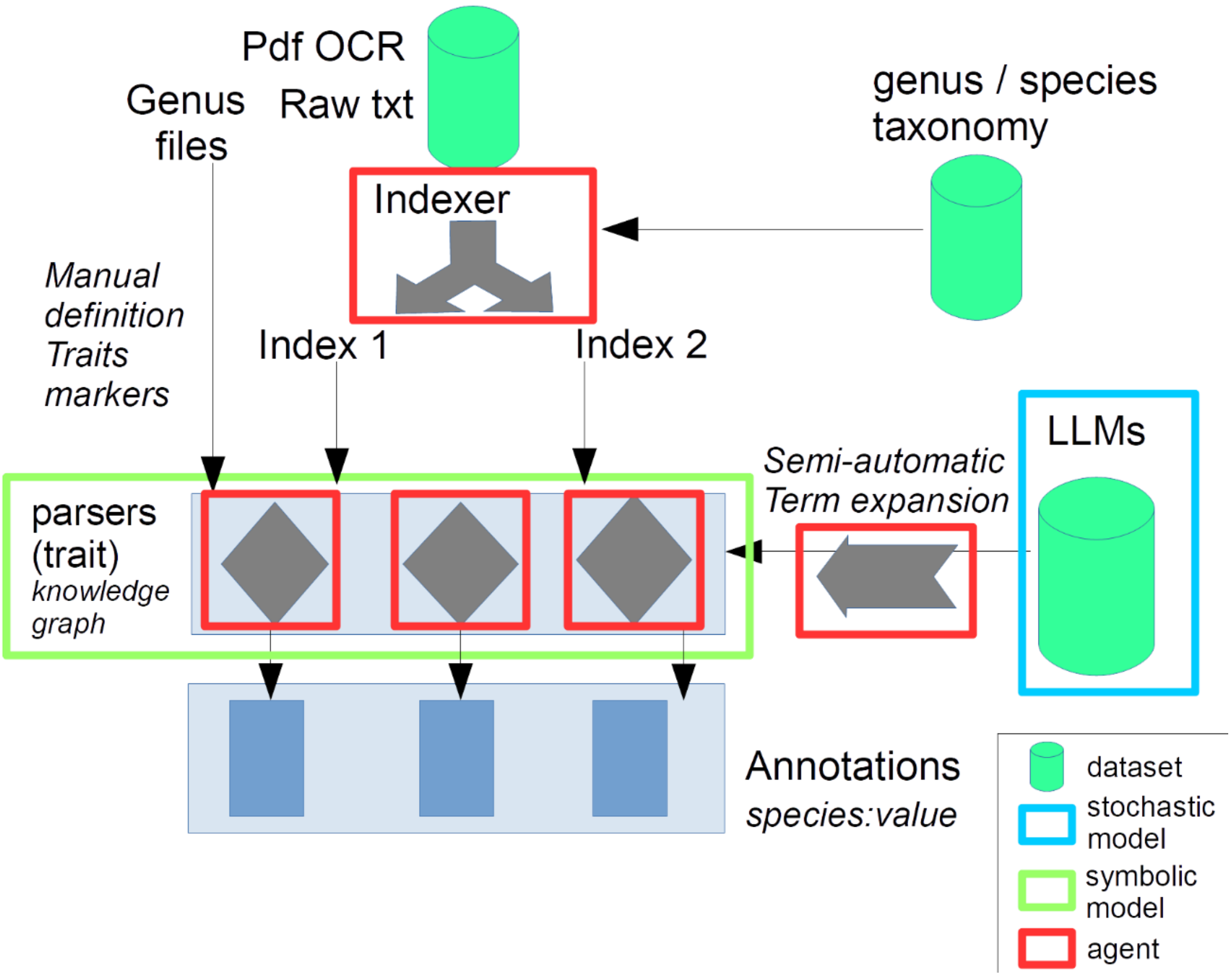

**Fig. 2** Pipeline illustrating the processing of documents from PDF scanning to species-specific annotation

The workflow consists of five stages. The first stage performs OCR on PDFs to convert scanned content into machine-readable text. The second stage handles indexing, clustering page content into unambiguous file groups and identifying passages relevant to specific categories, ultimately generating passage embeddings. The third stage defines initial rule sets for classification. The fourth stage enriches these rules using an ensemble of LLMs, leveraging their reasoning capabilities. The fifth stage applies the refined rules across all indexed files to produce structured annotations. Each stage corresponds to a set of agents, with subsequent stages dependent on the outputs of preceding ones.

### 4.2 Species Name Recognition and Normalization

The OCR process was described in the previous section (cf. Dataset section). Accurate recognition of species names is crucial for both the OCR and indexing steps. To address this, we developed *WordGen*, a name-matching algorithm based on transformation rules, including insertion, deletion, and substitution. The algorithm is designed to tolerate minor typographical or OCR-induced errors, particularly in species names. *WordGen* accepts either a genus or a genus + epithet as input and performs approximate string matching against a curated botanical dictionary.

As shown in Figure 3, the algorithm performs fuzzy matching of species names by comparing a given input (e.g., "leptostylis gzrandidolia") against a curated dictionary of epithets. The figure illustrates one example rule among several applied to the input. The algorithm extracts valid species epithets associated with the genus Leptostylis and generates candidate corrections using insertion rules (e.g., inserting a single letter at each possible position). Candidates are then validated by checking for matches against the input according to a defined pattern, ultimately returning the corrected name "Leptostylis grandidolia".

The function first cleans and splits a botanical name into genus and species components, then checks for exact matches. If no exact match is found, it searches for approximate matches by simulating letter substitutions (rule 1), insertions (rule 2), and deletions (rule 3). When approximate matching fails, the function attempts loose prefix matching (rule 4). Finally, it flattens and sorts the results, returning all possible matching names (see Figures 3 and 4).

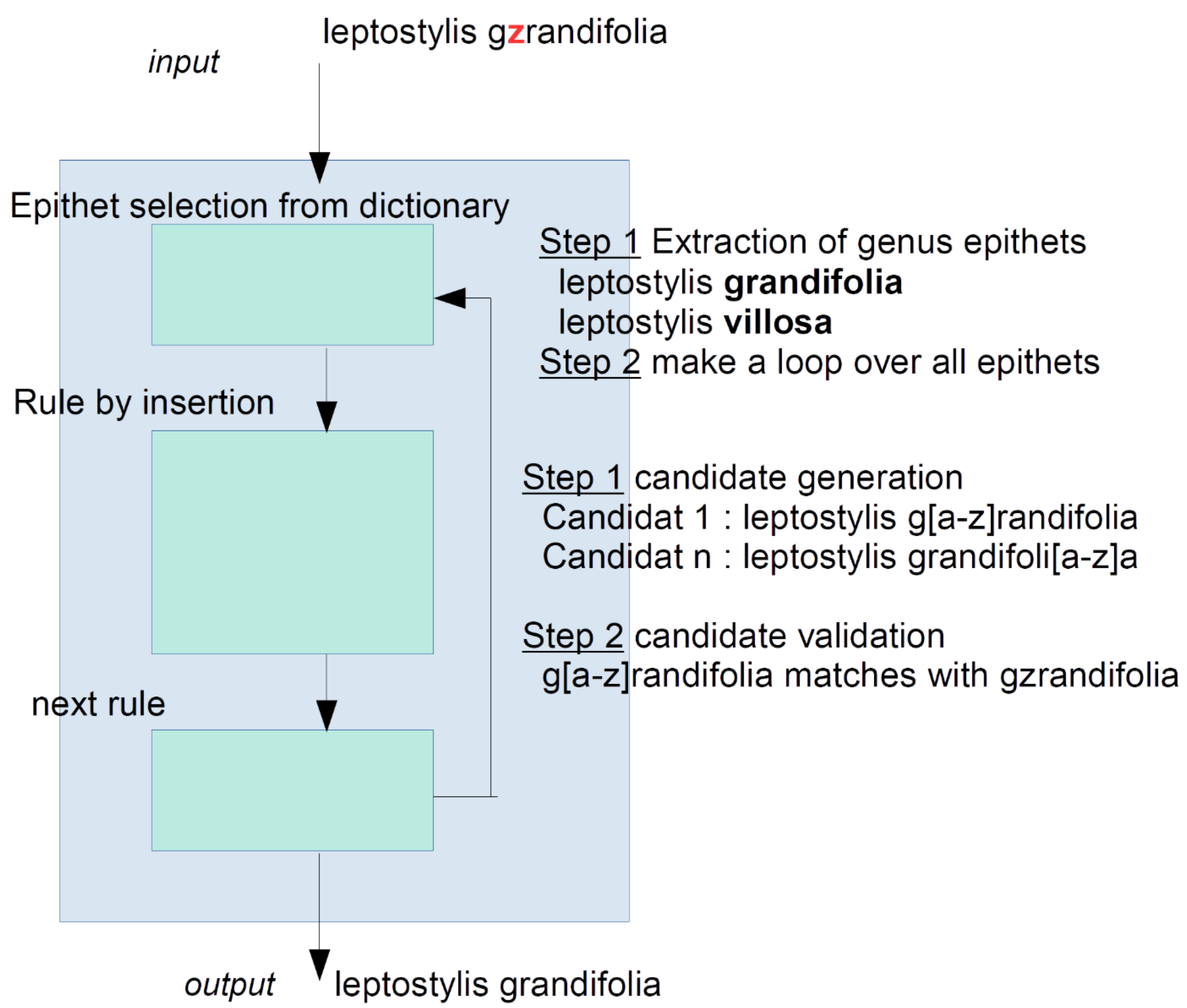

**Fig. 3** WordGen workflow

```
Input  : name (string), dict (dictionary)
Output : results (list)

Step 1 – Split name into genus and species

Step 2 – Direct exact match check in dict
    – If genus exists in DicGenus and species exists in its keys → return [species]
    – If only genus given and exists in DicGenus → return [genus]

Step 3 – Fuzzy matching candidates from entry = [species] or [genus]

-For each item in entry

  3.1 – Substitution pattern
    – For each character position in w:
      • Replace that character with \w wildcard
      • Create regex pattern and collect matches in genus keys (or all genera)

  3.2 – Insertion pattern
    – For each position (including start and end):
      • Insert \w at that position
      • Create regex pattern and collect matches with dict

  3.3 – Deletion pattern
    – For each character position:
      • Remove that character
      • Create regex pattern and collect matches with dict

  3.4 – Loose prefix match (only if no matches so far)
    – Take first 4 letters of w
    – Regex: "^prew[-\w]*$" against genus keys with dict
    – Keep results where length difference < 7 and w length > 4

Step 4 – Flatten and sort results
    – Flatten nested lists
    – Remove duplicates
    – Sort alphabetically
```

**Fig. 4** *WordGen* pseudocode. The input can be either a genus or a genus + epithet

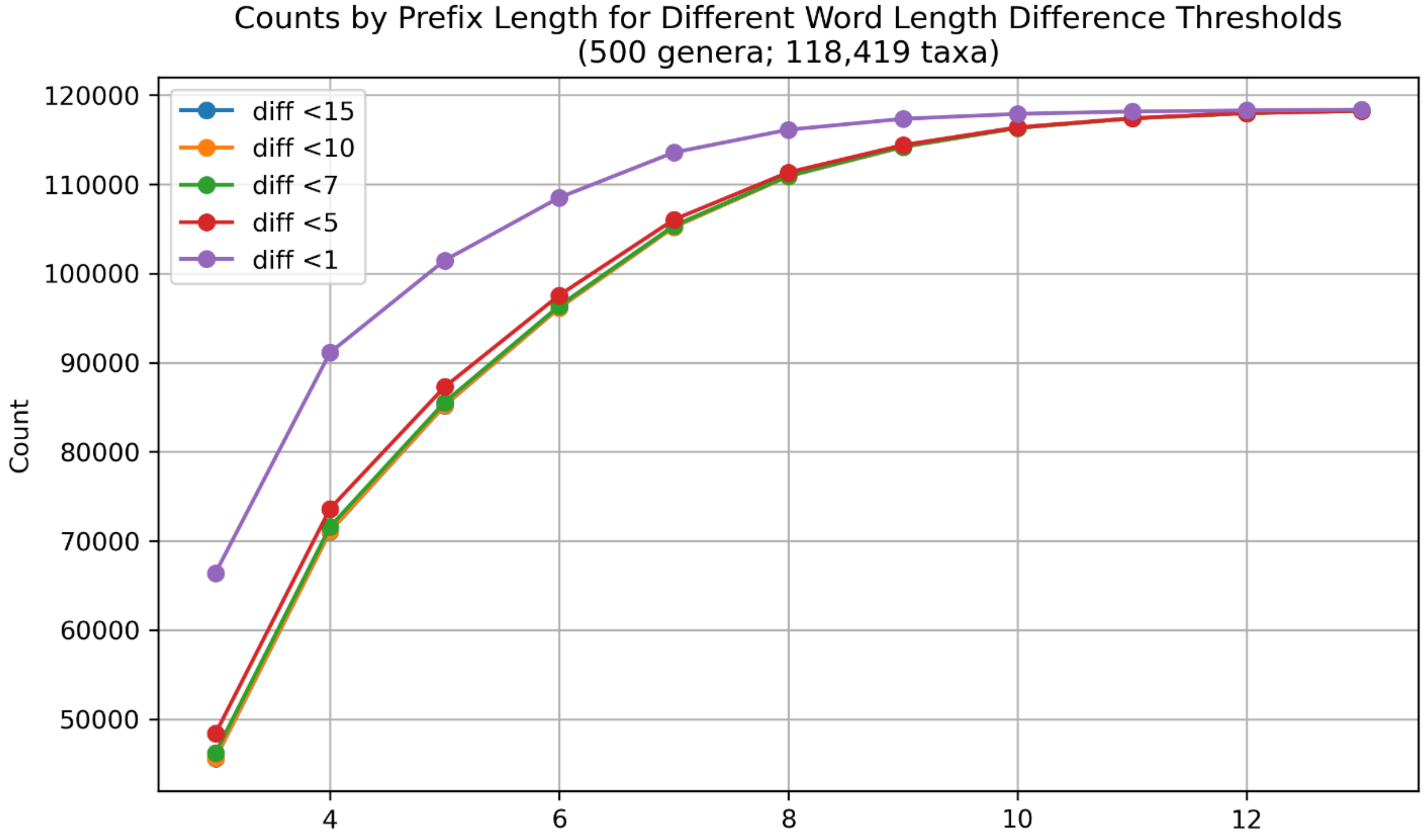

**Fig. 5** Accuracy of detecting correct matches as a function of prefix size and word-length difference

We tested the prefix rule on 500 genera and their 118,000 species epithets. For each species name, the rule was applied by truncating the species epithet to prefixes ranging from the first 3 to the first 13 characters and checking for a match in the dictionary. Cases with multiple matches were considered ambiguous and counted as potential errors, while the absence of a match was also treated as an error. Candidate matches were further filtered based on the length difference between the predicted epithet and the true epithet. For instance, if the true epithet is novoides and the rule returns novocaledonia, the length difference is 5; allowing a maximum difference of 3 characters would exclude novocaledonia as a valid candidate. Figure 5 shows that exact matches with longer prefixes yield high accuracy, but at the cost of generalization. Prefix lengths of 5–6 characters provide a balance between accuracy and generalization, with length differences of 5–7 characters performing well. Based on these results, we selected a 5-character prefix and a maximum length difference of 7.

### 4.3 Text Segmentation and Passage Embedding (Indexing)

We process the OCR text corpus in two stages. First, each file is segmented into sections corresponding to a specific genus. A single genus can include several thousand species (taxa), and each species name consists of a genus name followed by a specific epithet. Because epithets can be shared across different genera, this introduces significant ambiguity. By dividing each text file into genus-specific segments, we reduce uncertainty and improve accuracy in species name identification.

The second stage involves identifying segments within each genus-specific file that describe the taxonomic features of individual species. We refer to this process as indexing, which entails extracting the first and last lines of every passage in the genus file where a given species is mentioned and described. A single species may have multiple corresponding passages.

*Segmentation (first stage)*

We begin by splitting the text into individual lines and checking each line for the presence of a genus name based on a predefined rule. We then construct a block by copying each line into a file named after the current genus (e.g., Genus A) until the next genus (e.g., Genus B) is detected.

This process can introduce biases: if Genus B is not detected, its descriptions may be incorrectly appended to the Genus A file. Conversely, if Genus B is a false positive, the lines stored in Genus B.txt actually belong to Genus A.txt

The procedure is summarized in Figure 6 (pseudo-code). First, the text is cleaned by removing surrounding punctuation, including characters such as , ; : ( ) \n ' . = * [ ] ~ / \. Certain gendered words—such as paris, fine, and costa—that may interfere with accurate segmentation are also eliminated.

Next, for words not preceded by a number (a common OCR artifact) and longer than five characters, we test whether a generated pattern matches an entry in the dictionary. Patterns are created by inserting wildcard characters into the word (e.g., pychandra becomes the masked form pyc.andra), enabling detection of the correct dictionary term (e.g., pycnandra).

We then apply the following rules:

- Rule 1: Within a line, detect an uppercase expression present in the genus dictionary followed by a number and a period.

· Rule 2: Within a line, detect an uppercase expression present in the genus dictionary where the expression length exceeds three characters, appears within the first three positions of the line, and the total line length is fewer than five words.

Exception rules are added to handle "orphan" species—species names appearing in a standalone paragraph not clearly associated with a genus section. In such cases, the species description is detected and copied to the appropriate genus file.

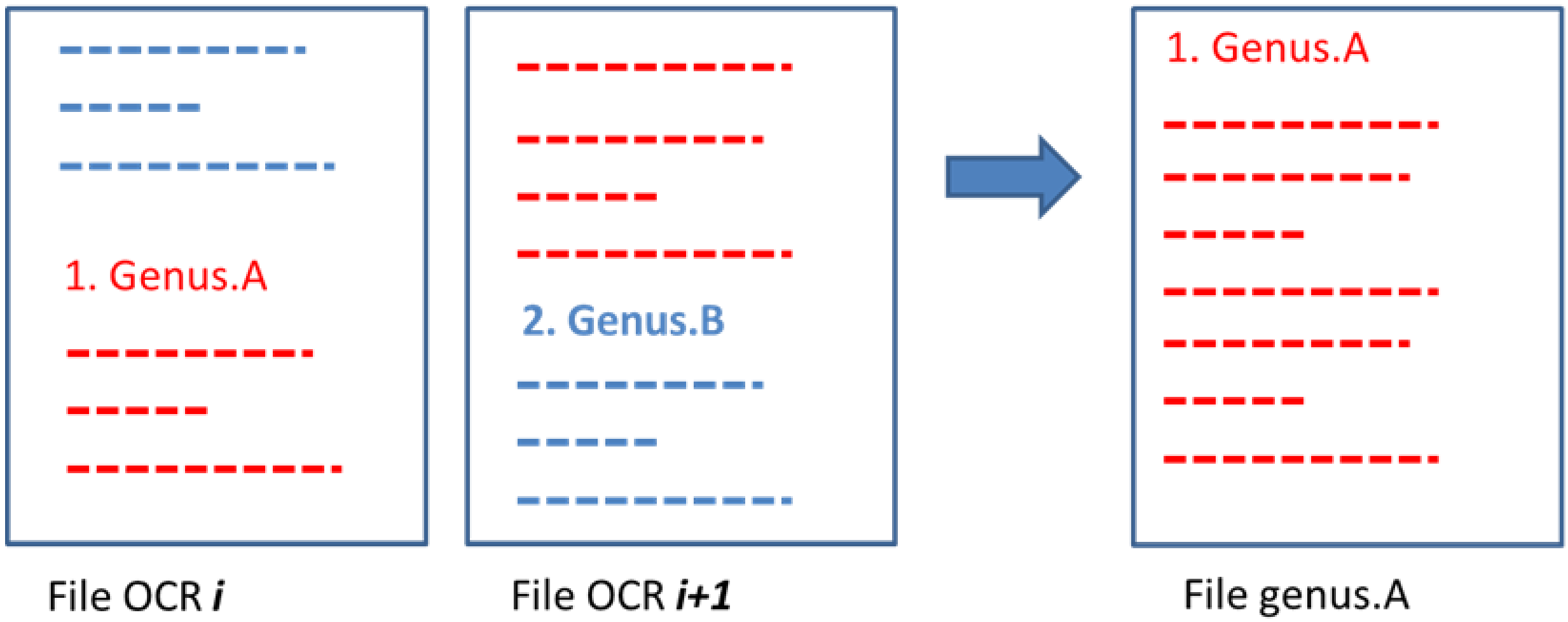

**Fig. 5** Distribution of genera across pages

```
Input list of files {f_i}, Dictionary D

For each f_i
    {L_i} = read lines of f_i

For each L_i
    {W_j}= split(L_i)  #preprocessing1
    ExceptionList: "paris", "papo", "lens", "lablab","lass" ,"fine", "costa","anno"
       ,"rain"  W_j not in D #preprocessing2
    if( len(W_j)>5 and W_j not in D ): #preprocessing3
       W_m = mask one character any position
       result = find(W_m, L_i)
       If W_m in D  W_j = W_m
    if   W_j in D and W_j  is Uppercase and "[0-9]." in L_i )
       MainGenus = W_j   #rule1
    if   W_j in D and W_j  is Uppercase and len(W_j)>3 and pos(W_j) <3 and len({W_j})<=5
       MainGenus = W_j   #rule2

write(L_i)  in   "MainGenus.txt" file
```

**Fig. 6** Pseudocode for text segmentation

In summary, a genus file contains all descriptions related to the species within that genus. Typically, it corresponds to a single section, but in some cases, it aggregates multiple parts of a volume. Consequently, the description of a species may appear in different segments of the text.

*Passage embedding (second stage)*

The objective of the index structure is to map each file path and collection to its corresponding genus, which in turn lists species names along with their start and end positions. Since a species name can appear in multiple locations within a file, it is necessary to index the species name separately. For a given genus file, the goal is to identify a sequence of blocks in the format: *name_index : begin_line : end_line*.

The algorithm proceeds as follows. Each line is preprocessed by checking for patterns present in a dictionary. Patterns are generated using a masked language model, where selected letters are replaced with wildcard characters. For example, “puchandra” is transformed into the masked pattern “p.c.andra” (. is a mask and can be any character) enabling the detection of similar terms such as “pycnandra” in the dictionary. This procedure is applied to both genus and species names. Punctuation surrounding words—including , ; : ( ) \n ' . = * [ ] ~ \ / “ ‘ _—is removed prior to matching.

Next, rule-based checks are applied. Rule 1 verifies the presence of the genus within the first four words of a line. Rule 2 checks for the co-occurrence of the genus and its epithet within the same line.

```
Input list of genre files {f_i} for a specific collection , Dictionary D

•for each f_i
•  LineNumber=1 ; ListSpecies{}; ListSpeciesIndex{}
•  G = split(path)                                  #get genre name
•  {L_i} = read lines of f_i
•  {S_i} = read species of G from D
•  {S_m} = mask two characters any position for each S_i         #preprocessing1
•  G_m = mask two characters any position for G in one regular expression

•for each L_i
•  {W_j}= split(L_i)                                              #preprocessing2
•  if  find(G_m, L_i)  and pos < 4                           #rule1
•    for each S_m
•      if find(S_m, L_i)   ListSpecies ← (S_m , LineNumber)       #rule2
•for each Species_i in ListSpecies
•  LineNumberBegin ←  ListSpecies ← (Species_i   , LineNumber)
•  LineNumberEnd    ←  ListSpecies ← (Species_i+1 , LineNumber)
•  ListSpeciesIndex ← (Species_i , LineNumberBegin , LineNumberEnd)

•write(ListSpeciesIndex)  in  "index_collection.txt" file
```

**Fig. 7** Pseudo-code of indexing

### 4.4 Rule-based Knowledge Extraction from textual taxonomic description

The objective of this study is to automate the extraction of trait values from textual taxonomic descriptions across multiple species. Here, a trait is defined as a property of a living organism, such as a plant, comprising a concept and an associated value. For example, in the trait "Leaf Shape," the concept is "shape" while "denticulate" represents the specific value corresponding to this concept. Traits can be further categorized into types such as ecological, agricultural, or taxonomic, enabling structured representation and downstream computational analysis.

Our goal is to extract comprehensive descriptions of plant characteristics, which are essential for global plant databases and for supporting botanical research. These descriptions can be derived from both images and taxonomic texts. Botanists and ecologists typically characterize plants through phenotypes, expressed as traits. From textual sources, we can extract morphological traits such as leaf margin shape or the arrangement of flowers, capturing whether they are solitary or clustered, enabling structured representation for computational analysis.

According to the FLOPO ontology (Hoehndorf et al., 2016), there are approximately 24,000 plant traits globally, including associated trait-value distributions. To facilitate automated trait extraction, we develop rules derived from collocated markers identified within the corpus. This process involves three steps: (1) identifying markers in context, (2) defining extraction patterns based on these markers, and (3) evaluating the resulting rules for accuracy and coverage.

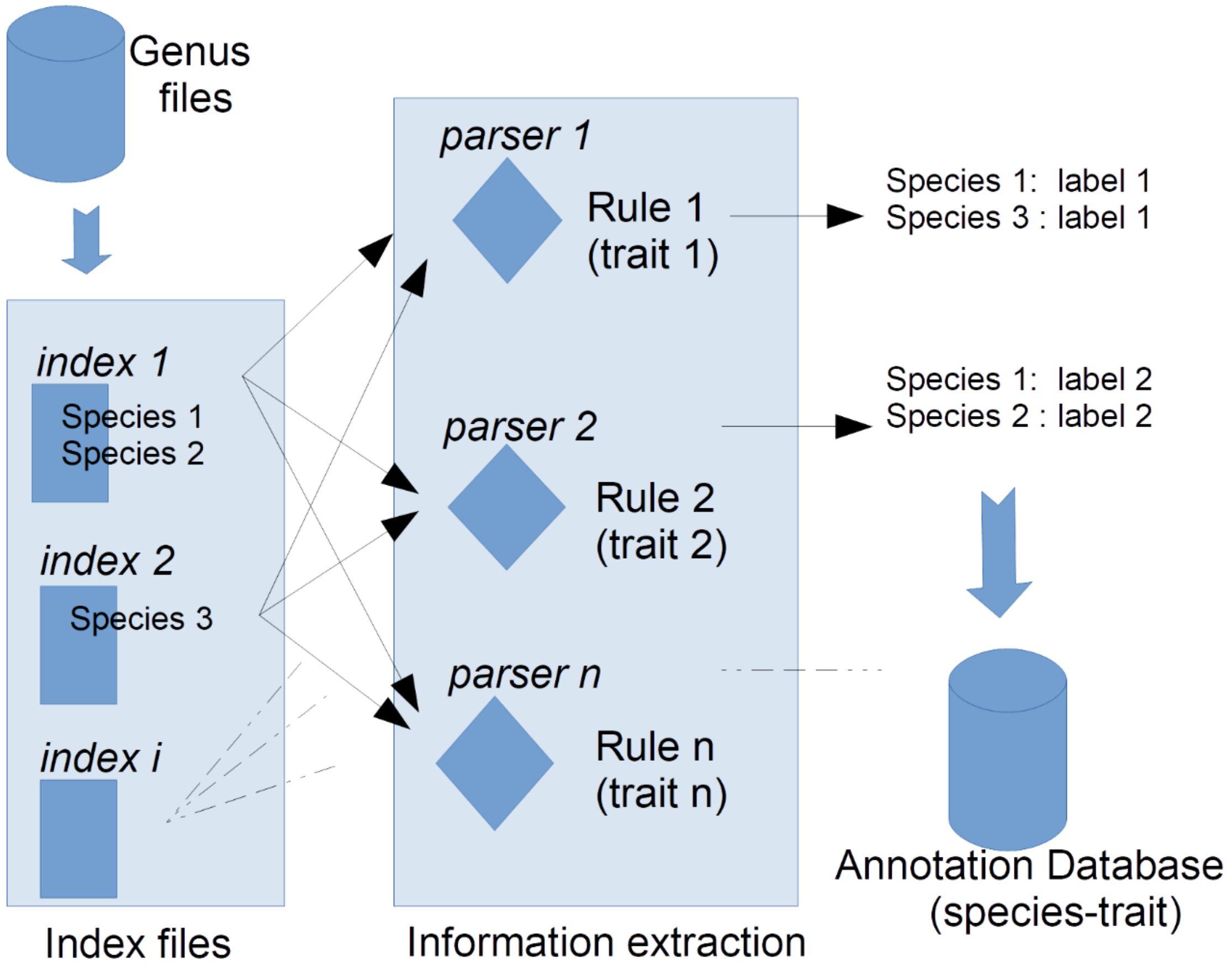

**Fig. 8** Pipeline of rule application. Each parser operates as an independent agent, with the number of parsers equal to the number of rules (#parsers = #rules)

For example, to extract information about leaf size, we first analyze contexts in which leaves are described, identifying potential markers such as *leaf* or *limb* (from the French *feuille* or *limbe*). We then focus on dimensional expressions that typically indicate size, such as “4 cm” or “4–10 cm”. Based on these observations, we define a rule that detects the co-occurrence of:

- a primary marker, such as *leaf/leaves* or *limb/limbs* (captured using the regular expression *feuille[s]?|limbe[s]?*), and

- a secondary size marker, represented by a numerical value or range followed by a unit (e.g., 4 cm or 4–16 cm), using a regular expression such as *[0-9]+ ?(cm|mm)|[0-9]+[-][0-9]+ ?(cm|mm)*.

**Tab. 2** Overview of traits used to define extraction rules

| | Traits (1-14) | | Traits (15-28) |
|---|---|---|---|
| 1 | Stem texture | 15 | Leaflet/leaf base – acute |
| 2 | Presence of thorns | 16 | Leaflet/leaf base – rounded |
| 3 | Leaf margin | 17 | Leaflet/leaf base – auriculate |
| 4 | Leaflet/leaf size | 18 | Leaflet/leaf base – I don't know |
| 5 | General shape of the lamina – elliptic | 19 | Leaf attachment on the stem |
| 6 | General shape of the lamina – oval | 20 | Do you observe any hairiness (pubescence) on the lamina? |
| 7 | General shape of the lamina – obovate | 21 | Presence of flowers |
| 8 | General shape of the lamina – lanceolate | 22 | Presence of fruit |
| 9 | General shape of the lamina – no leaf | 23 | Flowers and fruits not visible |
| 10 | Leaf arrangement on the stem | 24 | Inflorescence |
| 11 | Leaflet/leaf tip – acute | 25 | Infructescence |
| 12 | Leaflet/leaf tip – rounded | 26 | Neither inflorescence nor infructescence |
| 13 | Leaflet/leaf tip – auriculate | 27 | Simple flower |
| 14 | Leaflet/leaf tip – I don't know | 28 | Simple fruit |

For example, the phrase "Leaves narrowly obtriangular-obovate, 4–16 cm long" can be successfully extracted because it satisfies both the primary and secondary conditions defined by the regular expressions.

Figure 8 illustrates the overall pipeline, from PDF processing to rule application and trait-value extraction. For rule definition, we focused on a shortlist of 28 morphological traits (see Table 3). Table 4 presents examples of species descriptions containing features corresponding to these traits; only three examples are shown, with the complete list provided in the supplementary information.

**Tab. 3** Traits and their associated extraction rules (markers). The first column specifies the trait name, the second column indicates the primary marker, and the third column provides the secondary marker, if applicable. The final column reports the number of species matches identified for each trait

| | Trait | First marker | Second marker | #taxa |
|---|---|---|---|---|
| 1 | Stem texture | stem \| main axis | glabrous \| erect \| ascending \| robust \| prostrate \| creeping \| twining \| climbing | 225 |
| 2 | Presence of spines | spine \| spiny \| spiny \| microspine | NULL | 94 |
| 3 | Leaf margin | leaf \| margin \| leaflet | serrulate \| tooth \| serrate \| spinescent \| ciliate \| denticulate \| crenate \| crenellate \| incised \| spreading \| sinuate \| eroded \| lobed \| wavy \| toothed \| dentate \| runcinate | 420 |

### 4.5 Rule definition based on an ensemble of LLMs

As described in the previous section (cf. rule-based knowledge extraction), we manually define rules to annotate plant species with morphological traits. However, the domain-specific vocabulary is extensive and can only be fully handled by experts. We leverage LLMs as virtual domain experts to enrich these rules with new terms, thereby increasing annotation coverage. To further enhance term discovery, we implement term expansion (Turenne et al., 2025c) using an ensemble of LLMs within an iterative algorithm we call Object Class Learning (OCL). Figure 4 illustrates the overall pipeline integrating the LLM. While other frameworks employ alternative LLM strategies, such as pllama (pllama, 2024), which fine-tunes models on plant-related literature, our approach preserves rule-based extraction to maintain a high level of explainability and accuracy.

One of the key methods in this context is ontology learning, which aims to automatically or semi-automatically construct formal knowledge representations in the form of ontologies. An ontology models a domain through concepts, their properties, and the relationships between them. The simplest and most common representation in this framework is the RDF triples (Resource Description Framework), where knowledge is expressed as subject-predicate-object triples forming a semantic graph. For example, in the sentence "This stem is ascending" one can extract the triple:

- Subject: stem (an instance or concept of the class stem)
- Predicate: is (a membership or property relation)
- Object: ascending (an instance or concept of the class type of stem)

The algorithm proceeds in iterative rounds with two main steps:

Step 1: Enrich the vehicle class by extracting new vehicle instances associated with known energy class terms (e.g., electric, diesel, kerosen). From this, additional vehicles such as sailboat and barge are discovered.

Step 2: Using the expanded set of stem (stem, main axis), find new energy class terms linked to these vehicles. This leads to the discovery of stem types like erect prostrate, creeping, and twining.

In the following example, we apply the OCL algorithm to the concept of stem texture by linking ObjectClassLeft = "stem" with ObjectClassRight = "type of stem of a plant" during Round 1. This iteration also employs representative example sentences demonstrating how the orientation or growth habit of a plant's stem or main axis is described:

```
example1 = "The stem is ascending"
example2 = "The main axis is ascending"
example3 = "The stem is erect"
example4 = "The main axis is erect"
example5 = "The stem is prostrate"
example6 = "The main axis is prostrate"
example7 = "The stem is creeping"
example8 = "The main axis is creeping"
example9 = "The stem is twining"
example10 = "The main axis is twining"
```

These examples illustrate the variation in terminology used to describe similar morphological traits and serve as input for rule-based or LLM-assisted annotation.

The results were generated using GPT-4 with a temperature setting (T) of 0.4 and a maximum token length (MT) of 100, during Round 15 of the evaluation process.

After providing the 10 examples, the system produced 114 candidate terms for term expansion, including:

['Atactostele', 'Axillary bud', 'Bulb', 'Cane', 'Caudex', 'Caulome', 'Culm', 'Cladode', 'Collar', 'Corm', 'Dictyostele', 'Sucker (shoot)', 'Internode', 'Eustele', 'Fasciation', 'Fascicle', 'Water sprout', 'Halm', 'Rush', 'Laticifer', 'Free', 'Palm', 'Phylloclade', 'Pleionome', 'Protostele', 'Pseudobulb', 'Peduncle', 'Taproot', 'Creeping root', 'Radial', 'Fibrous rootlet', 'Branch', 'Twig', 'Offset', 'Rhizome', 'Rhytidome', 'Rosette', 'Runner', 'Scape', 'Scion', 'Sympodium', 'Siphonostele', 'Stolon', 'Acaulescent stem', 'Winged stem', 'Honeycombed stem', 'Ringed stem', 'Tree-like stem', 'Jointed stem', 'Ascending stem', 'Modified aerial stem', 'Forked stem', 'Grooved stem', 'Composite stem', 'Prostrate stem', 'Hollow stem', 'Cylindrical stem', 'Erect stem', 'Decumbent stem', 'Thread-like stem', 'Flexuous stem', 'Floral stem', 'Flower-bearing stem', 'Floating stem', 'Wilted stem', 'Foliary stem', 'Fruit-bearing stem', 'Glabrous stem', 'Climbing stem', 'Herbaceous stem', 'Woody stem', 'Smooth stem', 'Knotted stem', 'Knobby stem', 'Solid stem', 'Hairy stem', 'Quadrangular stem', 'Branched stem', 'Creeping stem', 'Drooping stem', 'Beaked stem', 'Rough stem', 'Reticulated stem', 'Simple stem', 'Sinuous stem', 'Modified underground stem', 'Spinescent stem', 'Spiral stem', 'Submerged stem', 'Succulent stem', 'Twisted stem', 'Vertical stem', 'Sticky stem', 'Twining stem', 'Scaled stem', 'Emergent stem', 'Spiny stem', 'Upright stem', 'Tiered stem', 'Flaring stem', 'Tiller', 'Trunk', 'Tuber', 'Turion', 'Prostrate', 'Erect', 'Ascending', 'Creeping', 'Twining', 'Spike']

Real term expansion is obtained using OCL with simultaneously LLAMA, GPT and MISTRAL. We run OCL three times on same LLM. Then this is the term expansion we summarize by this short list .

- *stem*: 'branch', 'twig', 'rhizome', 'stem (caulis)', 'scape'
- *type plant stem*: 'trailing', 'underground', 'drooping', 'vertical', 'floating', 'submerged', 'erect', 'filiform', 'orthotropic', 'plagiotropic', 'sarmentose', 'twisted', 'flared', 'decumbent', 'cespitose', 'tracing', 'flexuous', 'hanging', 'spiraled', 'caulescent'

## 5 Evaluation and discussion

### 5.1 Experimental Environment

We utilized two primary APIs for accessing large language models: OpenAI's GPT series via the OpenAI platform, and Meta's LLaMA and Mistral models via the LLaMA API. All models employed are instruction-tuned, providing cloud-based access to high-capacity language models without the need for local deployment. This setup offers advantages in integration and scalability, though it introduces cost constraints, as usage is generally billed per token or per API call.

Each model exposes tunable hyperparameters that influence text generation. The *temperature* (T) controls randomness, ranging from 0.0 (fully deterministic) to 2.0 (highly stochastic). The *maximum length* (ML) sets the upper limit on the number of tokens generated in a single response, typically capped at 4096 tokens. The *top-p* parameter implements nucleus sampling, restricting generation to tokens within a cumulative probability mass—for example, *top_p* = 0.5 selects from the top 50% most probable next tokens.

### 5.2 Evaluation Measures

We conducted the evaluation using our botanical dataset as the framework. Two primary metrics were considered: Number of Trait Extractions and Average Number of Correct Annotations per Species.

Number of Trait Extractions is computed both per trait—counting how many times each specific trait is successfully extracted across all species—and overall, representing the total number of trait extractions across all traits and species. This measure also enables quantification of coverage and identification of traits that are more or less successfully extracted.

Average Number of Correct Annotations per Species evaluates the average number of traits correctly annotated for each species, reflecting the accuracy and completeness of trait annotation at the species level. It is computed as:

$$Average\ annotations/species\ = \frac{Total\ correct\ trait\ annotations\ across\ all\ species}{Number\ of\ species}$$

**5.3 Single Monolithic LLM without Agents: Question Answering (QA)**

We designed a trait-based information extraction task by directly prompting an LLM with a species description text file. The instruction was: "You are a botany expert. Indicate whether an inflorescence is present. Indicate whether a fruit is present. Do not add any additional text."

In this task, the LLM was asked to identify the presence of an inflorescence and a fruit without generating any additional text. The results (Table 3) indicated an accuracy of 56% (5/9) for inflorescence detection and 67% (6/9) for fruit presence.

**Tab. 4** LLM responses for trait values in a species description file

| File | Ground Truth | LLM |
|---|---|---|
| FLORE_01-00000024.txt | inflorescence: no,<br>fruit: yes | yes<br>yes |
| FLORE_01-00000046.txt | inflorescence: no,<br>fruit: yes | yes<br>No |
| FLORE_01-00000050.txt | inflorescence: no,<br>fruit: yes | No<br>No |
| FLORE_01-00000061.txt | inflorescence: no,<br>fruit: yes | yes<br>yes |
| FLORE_02-00000039.txt | inflorescence: yes,<br>fruit: yes | yes<br>yes |
| FLORE_02-00000066.txt | inflorescence: yes,<br>fruit: yes | yes<br>yes |
| FLORE_02-00000071.txt | inflorescence: oui,<br>fruit: yes | yes<br>yes |
| FLORE_02-00000243.txt | inflorescence: no,<br>fruit: yes | yes<br>No |
| FLORE_14-00000174.txt | inflorescence: yes,<br>fruit: yes | yes<br>yes |

This observation is highly correlated with the findings of Bang (2024), who reported that LLM accuracy is around 60% even when addressing factual events, highlighting issues of inconsistency and hallucination. These results further support the development of agentic LLM systems tailored to specialized domains.

**5.4 Result using a Botanical Dataset**

We used three area document subsets (New Caledonia, Senegal, Cameroon collections). Using rule-based extraction, with 29 traits we obtained 55,737 trait annotations. On average, this corresponds to 9.1 annotations per species, with 4,961 species having at least five annotations.

The plot on figure 9 shows the distribution of annotated traits per species. Most species have only a few traits annotated, with a sharp decline after 5 traits. A small subset of species accumulates more traits, but numbers rapidly decrease beyond 10 traits, highlighting uneven annotation density across the dataset. No species has more than 21 annotations out of 29 possible, which is expected since certain traits are mutually exclusive. For example, if a species is annotated with the presence of an inflorescence (implying multiple flowers), the rule for a unique flower trait will necessarily return a null value.

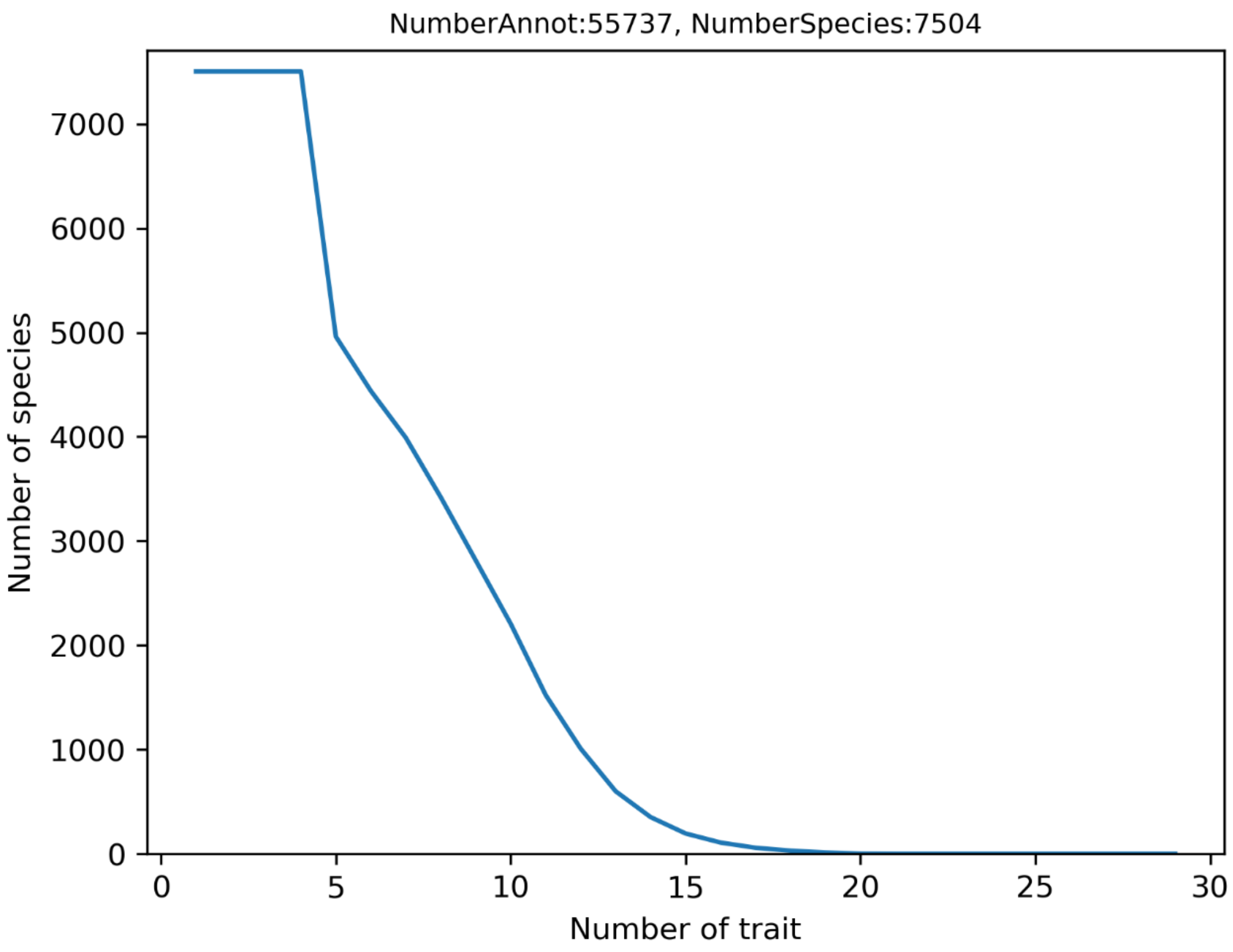

**Fig. 9** Number of species according number of annotations

### 5.5 Error propagation

Error propagation is assessed not at the output of an individual agent, but at the output of an agent that relies on the results of upstream agents. In our architecture, OCR processing can be performed using different algorithms: neural network–based methods, which are slower but more accurate, or geometry-based methods, which are faster but less precise. We compared *Tesseract* with *pdf2txt* and observed a slight improvement in species name identification (Turenne, 2025a). However, the overall number of annotations did not increase.

### 5.6 Ablation study

In our architecture, all agents are interdependent, except for the trait-value extraction agents (the "parser" agents). Removing an entire category of agents, such as OCR or indexing, is virtually impossible, as it would collapse the entire pipeline. For the parser agents, we compared performance with and without enrichment by an LLM-based agent. Vocabulary enrichment affected 75% of traits, yielding an average improvement of 59% in the number of annotations across all traits. However, the LLM agent relies on human intervention to select candidate terms for inclusion in the rules, as not all terms suggested by the iterative OCL algorithm can be directly integrated. While this partially limits automation, it enhances explainability and interpretability of the process.

## 6 Conclusion

We developed a knowledge extraction architecture for technical texts, in which extracted elements are enriched using a recurrence-based algorithm applied across a group of large language models. The extracted knowledge focuses on botany, with extraction rules specifically designed to identify information about plant morphological traits.

We present a modular, agent-based architecture for the large-scale extraction of morphological traits from botanical texts. The pipeline integrates OCR, segmentation, indexing, rule-based parsing, and vocabulary enrichment via large language models. Evaluation on regional botanical datasets shows that rule-based methods provide reliable trait annotation, while LLM-based enrichment significantly enhances coverage—affecting 75% of traits and increasing annotations by an average of 59%.

Although error propagation across agents and the reliance on human-guided rule refinement currently limit full automation, the system offers a transparent and explainable framework. By balancing rule-based precision with LLM-driven vocabulary expansion, this approach facilitates the construction of structured plant trait databases, providing valuable resources for taxonomy, ecology, and biodiversity research.

**Code and Data availability** Code will be available after acceptation online. Code is available on demand. For dataset see (Turenne et al, 2025a).

**Acknowledgements** The authors gratefully acknowledge IRD for providing access to the SEPHE computing cluster.

**Author contributions** All authors participated in discussions shaping the ideas, brainstorming the experiments, and defining the research questions in this study. NT conceived the idea for the study, developed the software for experiments evaluating the proposed approach as well as wrote the manuscript and prepared the figures and tables. EC and YS contributed to further refining the approach and experiment design as well as significantly improved the text, figures. JDZ provided supervised global project and provides criticisms. The manuscript was written with input from all authors.

**Funding** ANR EquipEx grant (21 ESRE 0053) supporting the project "e-COL+: Enhancement of Natural History Data in France" (May 2021 – April 2029), part of the "Equipements structurants pour la recherche" program under the third Investment for the Future Plan (PIA3).

**Supplementary information** file supp info

## Declarations

**Conflict of interest** The authors declare no Conflict of interest.